\pdfoutput=1

\documentclass[11pt]{article}

\usepackage{emnlp2021}

\usepackage{times}
\usepackage{latexsym}

\usepackage[T1]{fontenc}

\usepackage[utf8]{inputenc}

\usepackage{microtype}
\usepackage{graphicx}
\usepackage{amsmath}
\usepackage{booktabs}
\usepackage{url} 
\usepackage{xcolor}
\usepackage{latexsym}
\usepackage{amsfonts}
\usepackage{amsmath}
\usepackage{bbm} 
\usepackage{amssymb}
\usepackage{subfigure}
\usepackage{balance}
\usepackage{threeparttable}
\usepackage{multirow}
\usepackage{float}

\title{Is Multi-Hop Reasoning Really Explainable? \\ Towards Benchmarking Reasoning Interpretability}

  \newcommand*{\email}[1]{\texttt{#1}}

  \author{
    \textbf{Xin Lv}$^{1,2}$, \textbf{Yixin Cao}$^{3}$, \textbf{Lei Hou}$^{1,2}$\thanks{\quad Corresponding Author}\hspace{0.5em}, \textbf{Juanzi Li}$^{1,2}$ \\ \textbf{Zhiyuan Liu}$^{1,2}$,  \textbf{Yichi Zhang}$^{4}$, \textbf{Zelin Dai}$^{4}$\\
    $^1$Department of Computer Science and Technology, BNRist \\
    $^2$KIRC, Institute for Artificial Intelligence, Tsinghua University, Beijing 100084, China \\
    $^3$Nanyang Technological University, Singapore \\
    $^4$Alibaba Group, Hangzhou, China \\
    \email{lv-x18@mails.tsinghua.edu.cn, yixin.cao@ntu.edu.sg}\\
    \email{\{houlei,lijuanzi,liuzy\}@tsinghua.edu.cn}
    }

\begin{document}
\maketitle

\begin{abstract}
  Multi-hop reasoning has been widely studied in recent years to obtain more interpretable link prediction. However, we find in experiments that many paths given by these models are actually unreasonable, while little work has been done on interpretability evaluation for them. In this paper, we propose a unified framework to quantitatively evaluate the interpretability of multi-hop reasoning models so as to advance their development. In specific, we define three metrics, including path recall, local interpretability, and global interpretability for evaluation, and design an approximate strategy to calculate these metrics using the interpretability scores of rules. Furthermore, we manually annotate all possible rules and establish a \textbf{B}enchmark to detect the \textbf{I}nterpretability of \textbf{M}ulti-hop \textbf{R}easoning (BIMR). In experiments, we verify the effectiveness of our benchmark. Besides, we run nine representative baselines on our benchmark, and the experimental results show that the interpretability of current multi-hop reasoning models is less satisfactory and is 51.7\% lower than the upper bound given by our benchmark. Moreover, the rule-based models outperform the multi-hop reasoning models in terms of performance and interpretability, which points to a direction for future research, i.e., how to better incorporate rule information into the multi-hop reasoning model. Our codes and datasets can be obtained from \url{https:// github.com/THU-KEG/BIMR}.
\end{abstract}

\section{Introduction} 

\begin{figure}[t]
    \centering
    \includegraphics[width=\linewidth]{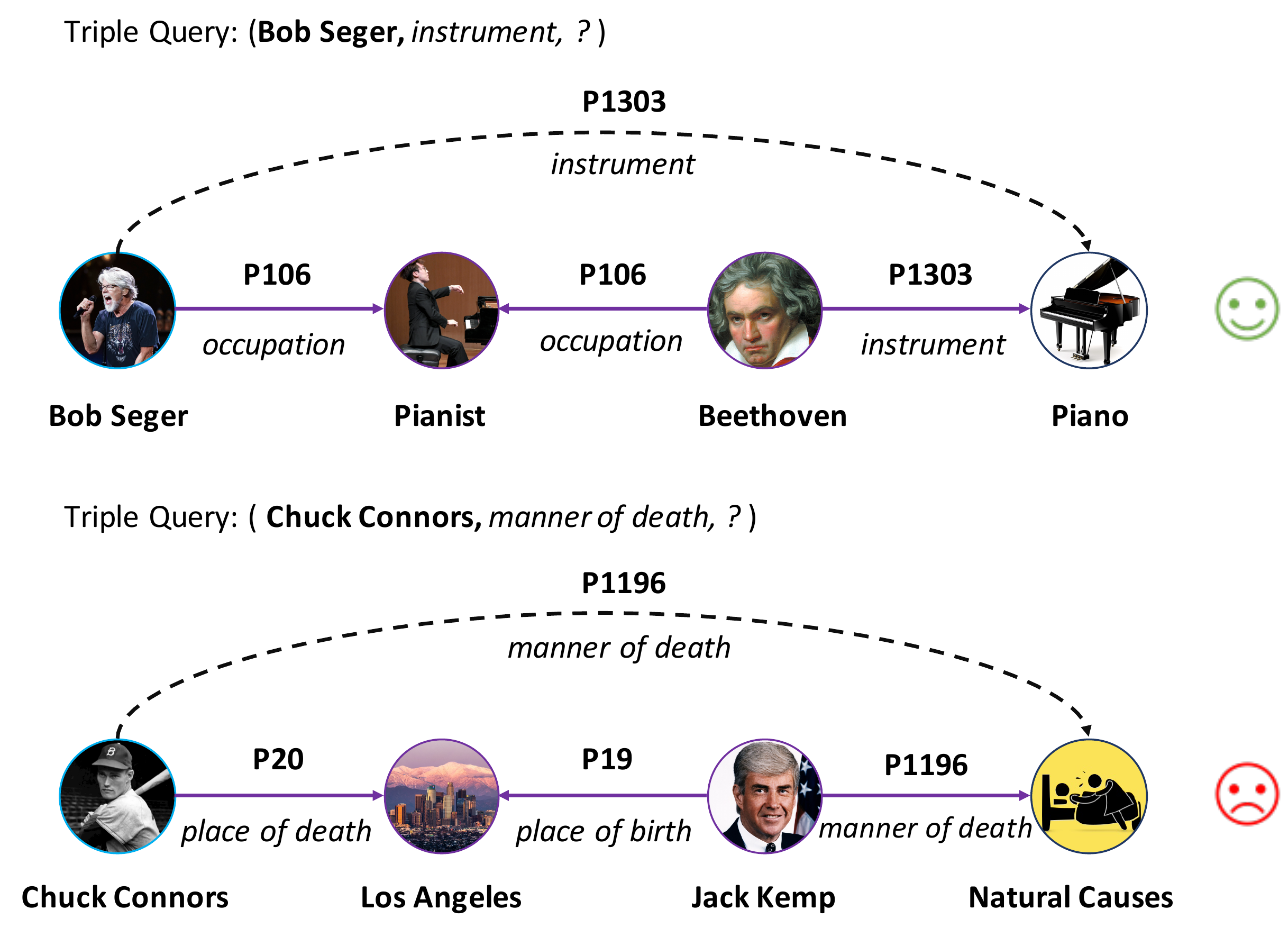}
    \caption{Illustration of reasoning paths for link prediction. Although both paths succeed in finding the correct tail entity, only the upper path is reasonable and the lower one is unreasonable.}
    \label{fig:intro}
  \end{figure}

Multi-hop reasoning for knowledge graphs (KGs) has been extensively studied in recent years. It not only infers new knowledge but also provides reasoning paths that can explain the prediction results and make the model trustable. For example, Figure~\ref{fig:intro} shows two inferred triples and their reasoning paths. Conventional KG embedding models (e.g., TransE~\cite{TransE}) implicitly find the target entity \textit{piano} given the query (\textit{Bob Seger}, \textit{instrument}, ?), while multi-hop reasoning models complete the triple and explicitly output the reasoning path (the purple solid arrows). Hence, multi-hop reasoning is expected to be more reliable in real systems, as we can safely add an inferred triple to KG by justifying if the path is reasonable.

Most existing multi-hop reasoning models assume that the output paths are reasonable and put much attention on the performance of link prediction. 
For example, MultiHop \cite{MultiHop} uses reinforcement learning to train an agent to search over knowledge graphs. The path found by the agent is considered as a reasonable explanation for the predicted result.
However, after manually labeling, we find that more than 60\% of the paths are unreasonable. 
As shown in the lower part of Figure \ref{fig:intro}, given a triple query (\textit{Chuck Connors}, \textit{manner of death}, ?), the multi-hop reasoning model finds the correct tail entity \textit{Natural Causes} via a reasoning path (the purple solid arrows). Although the model completes the missing triple correctly, this reasoning path is questionable, as the manner of death is not related to one's birth or death city. The reason for failed interpretability is mainly because many people are born and died in the same place \textit{Los Angeles}, and natural causes are the dominant death manner. Thus, the path is only statistically related to the query triple and fails to provide interpretability.
In experiments, we find that such unreasonable paths are ubiquitous in multi-hop reasoning models, suggesting an urgent need for an interpretability evaluation.

In this paper, we propose a unified framework to automatically evaluate the interpretability of multi-hop reasoning models. Different from previous works, which mostly rely on case study \cite{RLH} to showcase model interpretability, we aim at quantitative evaluation by calculating interpretability scores of all paths generated by models. In specific, we define three metrics: path recall, local interpretability, and global interpretability for evaluation (see Section \ref{sec:framework} for details). 
However, it is time-consuming to give each path an interpretability score since the number of possible paths that multi-hop reasoning can give is extremely large.
To address this issue, we propose an approximate strategy that abstracts reasoning paths into limited rules by ignoring entities with only relations left (see Equation \ref{eq:path} and \ref{eq:rule} for details). The total number of rules obtained in this way is much smaller than the number of paths, and we assign the interpretability score of the rule to its corresponding paths.

We explore two methods to give each rule an interpretability score, namely manual annotation and automatic generation by rule mining methods. The former is the focus of this paper. Specifically, we invite annotators to manually annotate interpretability scores for all possible rules to establish a manually-annotated benchmark (A-benchmark). This labeling process also faces a challenge, i.e., interpretability is highly subjective and hard to annotate. 
Different annotators may give various explanations. To reduce the variations, we provide the annotators with a number of interpretable options rather than asking them to give a direct score. Besides, for each sample, we ask ten annotators to annotate and take their average score as the final result. In addition to A-benchmark, we also provide benchmarks (R-benchmark) based on rule mining methods \cite{AnyBURL}. These benchmarks use the confidence of the mined rule as the rule's interpretability score. 
This approach is not as accurate as manual annotation but can be generalized to most KGs automatically.

In experiments, we verify the effectiveness of our benchmark BIMR.
Specifically, we obtain the interpretability of each model using the sampling annotation method and compare it with the results generated by our A-benchmark. The experimental results show that the gap between them is small, which indicates the approximate strategy has little effect on the results.
Furthermore,  we run nine representative baselines on our benchmarks. The experimental results show that the interpretability of existing multi-hop reasoning models is less  satisfactory and is still far from the upper bound given by our A-benchmark. Specifically, even the best multi-hop reasoning model is still 51.7\% lower in interpretability than the upper bound. This reminds us that in the study of multi-hop reasoning, we should not only care about performance but also about interpretability. Moreover, we find that the best rule-based reasoning method AnyBURL \cite{AnyBURL} significantly outperforms existing multi-hop reasoning models in terms of performance and interpretability, which points us to a possible future research direction, i.e., how to better incorporate rules into multi-hop reasoning.

\section{Related Work}

\subsection{Multi-Hop Reasoning}

Multi-hop reasoning models can give interpretable paths while performing triple completion. Most of the existing multi-hop reasoning models are based on the reinforcement learning (RL) framework. Among them, DeepPath \cite{DeepPath} is the first work to formally propose and solve the task of multi-hop reasoning using RL, which inspires much later work, e.g., DIVA \cite{DIVA}, and AttnPath \cite{ATTPath}. MINERVA \cite{MINERVA} is an end-to-end model with a wide impact that solves multi-hop reasoning task. On the basis of this model, M-Walk \cite{M-Walk} and MultiHop \cite{MultiHop} solve the problem of reward sparsity through off-policy learning and reward shaping, respectively. In addition, there are some other models such as the DIVINE \cite{DIVINE}, R2D2 \cite{R2D2}, RLH \cite{RLH} and RuleGuider \cite{RuleGuider} models that improve multi-hop reasoning from the four directions of imitation learning, debate dynamics, hierarchical RL, and rule guidance, respectively. CPL \cite{CPL} and DacKGR \cite{DacKGR} enhance the effect of models by adding additional triples to KG.

\begin{table}[t]
    \centering
    \small
    \scalebox{1.0}{
    \begin{tabular}{ccccc}
    \toprule[1pt]
    Dataset & \# Entity & \# Relation & \# Triple \\
    \midrule[0.5pt]
    WD15K & 15,817 & 182 & 176,524 \\
    \bottomrule[1pt]
    \end{tabular}
    }
    \caption{\label{tab:data_statistics} Statistics of WD15K. The three columns denote the number of entities, relations and triples, respectively.
    } 
  \end{table}

\subsection{Rule-based Reasoning}

Similar to multi-hop reasoning, rule-based reasoning can also perform interpretable triple completion, except that they give the corresponding rules instead of specific paths. Rule-based reasoning can be divided into two categories, namely, neural-based models and rule mining models. Among them, neural-based models \cite{NeuralLP,NTP,DRUM,CTPs} give the corresponding rules while performing triple completion, while rule mining models \cite{AMIE+,RLvLR,RuLES,AnyBURL} first mine the rules and then use them for completion.

\subsection{Interpretability Evaluation}

Few research work targets interpretability evaluation, although they admit the importance. 
Most multi-hop reasoning models rely on case study \cite{R2D2,RLH} to present the interpretability quality, while RuleGuider\cite{RuleGuider} samples tests and computes their differences for evaluation. There are some works in other areas \cite{gilpin2018explaining,yang2019bim,jhamtani2020learning} to test interpretability, but they cannot be directly applied to multi-hop reasoning tasks for knowledge graphs.

\section{Preliminary}

\noindent \textbf{Knowledge graph} (KG) is defined as a directed graph $\mathcal{KG} = \{\mathcal{E}, \mathcal{R}, \mathcal{T}\}$, where $\mathcal{E}$ is the set of entities, $\mathcal{R}$ is the set of relations and $\mathcal{T} = \{(h, r, t)\} \subseteq \mathcal{E} \times \mathcal{R} \times \mathcal{E}$ is the set of triples.

\noindent \textbf{Multi-hop reasoning} aims to complete KGs through interpretable link prediction. Formally, given a triple query $(h, r, ?)$, it needs to not only predict the correct tail entity $t$, but also give a path $(h, r, t) \leftarrow (h, r_1, e_1) \wedge (e_1, r_2, e_2) \wedge \cdots \wedge (e_{n-1}, r_{n}, t)$ as an explanation. 

\noindent \textbf{Rule-based reasoning} can be considered as generalized multi-hop reasoning and can also be evaluated on our benchmark. Given a triple query $(h, r, ?)$, it needs to predict the tail entity $t$ and give some Horn rules with confidence as an explanation, where the rule $f$ is of the following form:
\begin{equation}\small
  r(X, Y) \leftarrow r_{1}(X, A_1) \wedge \cdots \wedge r_{n}(A_{n - 1}, Y) .
\end{equation}
where capital letters denote variables, $r(...)$ is the head of the rule, the conjunction of $r_1(...), \cdots, r_n(...)$ is the body of the rule, and $r(h, r)$ is equivalent to a triple $(h, r, t)$.
In order to get the same path as the multi-hop reasoning task, we sort these rules in descending order of confidence and match them on KG.

\section{Benchmark}

In order to quantitatively evaluate the interpretability of multi-hop reasoning models, we first construct a dataset based on Wikidata (Section \ref{sec:dataset_construction}). After that, we propose a general evaluation framework (Section \ref{sec:framework}). Based on this framework, we apply an approximation strategy (Section \ref{sec:preparation_for_benchmark}) and build benchmarks with manual annotation (Section \ref{sec:benchmark_construction}) and mined rules (Section \ref{sec:benchmark_with_mined_rules}).

\subsection{Dataset Construction}
\label{sec:dataset_construction}

We curate an interpretable dataset \textbf{WD15K} based on Wikidata~\cite{Wikidata} as well as the widely-used FB15K-237~\cite{FB15K-237}. We aim to utilize the read friendly relations in Wikidata, meanwhile remain the entities from FB15K-237 unchanged. We rely on the Freebase ID property of each entity in Wikidata to bridge the two sources and the final statistics of our dataset WD15K are listed in Table \ref{tab:data_statistics}. We shuffle it and use 90\%/5\%/5\% as our training/validation/test set. Due to space limitations, we put the detailed steps of dataset construction in supplementary materials (Appendix A).

\subsection{Evaluation Framework}
\label{sec:framework}

We propose a general framework for quantitatively evaluating the interpretability of multi-hop reasoning models. Formally, each triple $(h, r, t)$ in the test set is converted into a triple query $(h, r, ?)$. The model is required to predict $t$ and possible reasoning paths. We thus compute an interpretability score for the model, which is defined based on three metrics: \textbf{Path Recall} (PR),  \textbf{Local Interpretability} (LI) and \textbf{Global Interpretability} (GI).

\textbf{Path Recall} (PR) represents the proportion of triples in the test set that can be found by the model with at least one path from the head entity to the tail entity. It is formally defined as
\begin{equation}
    \text{PR} = \frac{ \sum_{(h, r, t) \in \mathcal{T}^{\text{test}}} \text{Cnt}(h, r, t)}{|\mathcal{T}^{\text{test}}|} ,
\end{equation}
where $\text{Cnt}(h, r, t)$ is an indicator function to denote whether the model can find a path from $h$ to $t$. The function value is $1$ if at least one path can be found, otherwise, it is $0$. PR is necessary because, for most models, not every triple can be found a path from the head entity to the tail entity. For RL-based multi-hop reasoning models (e.g., MINERVA), the beam size is a critical hyper-parameter that has direct impacts on PR. The larger the beam size is, the more paths the model can find. In realistic, it, however, cannot be set to infinite. That is, there is an upper limit on the number of paths for each triple query $(h, r, ?)$. 
On the other hand, there may not be a path from $h$ to $t$, or we may not be able to match a real path on the KG for every rule. This leads to $\text{Cnt}(h, r, t)=0$.

\textbf{Local Interpretability} (LI) is used to evaluate the reasonableness of paths found by the model. It is defined as 
\begin{equation}
\label{equation:LI}
    \text{LI} = \frac{\sum_{(h, r, t) \in \mathcal{T}^{\text{test}}} \text{Cnt}(h, r, t) S(p)}{\sum_{(h, r, t) \in \mathcal{T}^{\text{test}}} \text{Cnt}(h, r, t)},
\end{equation}
where $p$ is the best path from $h$ to $t$ found by the model (the path with the highest score), and $S(p)$ is the interpretability score of this path which will be introduced in the following section.  

\textbf{Global Interpretability} (GI) evaluates the overall interpretability of the model, as LI can only express the reasonable degree of the path found by the model, but it fails to consider how many paths can be found. We define GI as follows:
\begin{equation}
    \text{GI} =  \text{PR} \cdot \text{LI}  .
\end{equation}
We summarize and compare LI and GI. Specifically, LI can reflect the interpretability of all paths that can be found, while GI evaluates the overall interpretability of the model.

\subsection{Approximate Interpretability Score}
\label{sec:preparation_for_benchmark}

Based on WD15K and the above evaluation framework, we can construct benchmarks to evaluate the interpretability quantitatively. However, $S(p)$ in the evaluation framework is difficult to obtain due to the huge number of paths. Thus, before the specific construction, some preparation work is needed, i.e., path collection and approximation strategy.

\paragraph{Path Collection.} This step aims to collect all possible paths from $h$ to $t$, so that our evaluation framework can cover the various outputs of multi-hop reasoning models. In specific, we first add reverse triple $(t, r^{-1}, h)$ for each triple $(h, r, t)$ in the training set. Then, for each test triple $(h, r, t)$ in WD15K, we use the breadth-first search (BFS) to search all paths from $h$ to $t$ on the training set within the length of 3, i.e., there are at most three relations in the path, which is widely used in multi-hop reasoning models (e.g., MultiHop). Because too many hops will greatly increase the search space and decrease the interpretability. After deduplication, we achieve the final path set $\mathcal{P}$ containing about 16 million paths, which covers all paths that may be discovered by multi-hop reasoning models. 

\paragraph{Approximation Optimization.} We propose an approximate strategy to avoid the impractical annotation or computation on the large set of paths $\mathcal{P}$ (i.e., 16 million). Based on observation, we find that the interpretability of a path mostly comes from the rules instead of specific entities. We thus abstract each path $p \in \mathcal{P}$ into its corresponding rule $f$, and use the rule interpretability score $S(f)$ as the path interpretability score $S(p)$, i.e., 
\begin{equation}
\label{eq:approx}
    S(p) \approx S(f) .
\end{equation}
Formally, for a path $p$
\begin{equation}
\label{eq:path}\small
    (h, r, t) \leftarrow (h, r_1, e_1) \wedge (e_1, r_2, e_2) \wedge (e_2, r_3, t) ,
\end{equation}
we convert it to the following rule $f$
\begin{equation}
\label{eq:rule}\small
    r(X, Y) \leftarrow r_{1}(X, A_1) \wedge r_{2}(A_1, A_2) \wedge r_{3}(A_2, Y) .
\end{equation}
After this conversion, we convert $\mathcal{P}$ into a set of rules $\mathcal{F}$. Because the rules are entity-independent, the size of $\mathcal{F}$ is greatly reduced to 96,019, and we only need to give each rule $f \in \mathcal{F}$ an interpretable score $S(f)$ to build the benchmark.

Next, we will introduce two types of methods to obtain the rule interpretability score.

\subsection{Benchmark with Manual Annotation}
\label{sec:benchmark_construction}

We manually label each rule in $\mathcal{F}$ with an interpretability score to form a manually-annotated benchmark (A-benchmark). The specific build process of this benchmark can be divided into two steps, namely pruning optimization and manual annotation. We will detail these two parts separately.

\paragraph{Pruning Optimization.} We propose a pruning strategy in order to save the annotation cost without causing a large impact on the final result.
Rule mining methods can automatically mine rules on the knowledge graph and give each obtained rule a confidence score.
Our pruning strategy is based on the assumption that those rules that are not in the list of rules mined by the rule mining methods, or those rules with very low confidence, have much lower interpretability scores. Below we show confirmatory experiments to verify our assumption.

In our specific implementation, we use AnyBURL \cite{AnyBURL} to mine rules on our training set and get 36,065 rules $\mathcal{F}^{A}$, where AnyBURL is one of the best rule mining methods, achieving the SOTA performance on many datasets. Based on rules in $\mathcal{F}^{A}$ and their confidence scores, we divide the entire rule set $\mathcal{F}$ into three groups, i.e., high confidence rules (H rules), low confidence rules (L rules), and other rules (O rules). We randomly sample 500 rules from each group and invite experts to manually label them with interpretability scores. The classification criteria of the rules and the labeling results are shown in Table \ref{tab:rule_pruning}. 

From Table \ref{tab:rule_pruning}, we can find that the average interpretability score of L rules is much lower than that of H rules. Although the average score of O rules is not as small as that of L rules, they correspond to too few paths (0.8M), while the number of rules is relatively large (75,027). Therefore, O rules can be considered as ``long-tail" rules. 

In order to save the cost of labeling, and as far as possible not to affect the final evaluation results, we only manually label the interpretability score of H rules. For the rest rules, we use the average score of this type of rule as their interpretability score. Specifically, referring to the results in Table \ref{tab:rule_pruning}, we uniformly consider the interpretability score as 0.005 for all L rules and 0.069 for all O rules

\begin{table}[t]
    \centering
    \small
    \scalebox{0.75}{
    \begin{tabular}{cccc}
    \toprule[1pt]
     & H Rules & L Rules & O Rules  \\
    \midrule[0.5pt]
    Criteria & $f \in \mathcal{F}^A \wedge c(f) \ge 0.01$ & $f \in \mathcal{F}^A \wedge c(f) < 0.01$ & $f \notin \mathcal{F}^A$ \\
    \# Rules & 15,458 & 5,534 & 75,027 \\
    \# Paths & 14.8M & 0.7M & 0.8M \\
    Avg Score & 0.216 & 0.005 & 0.069 \\
    \bottomrule[1pt]
    \end{tabular}
    }
    \caption{\label{tab:rule_pruning} Classification criteria and statistics of the three types of rules in $\mathcal{F}$, where $c(f)$ is the confidence score of rule $f$, \# Rules denotes the number of rules, \# Paths denotes the number of paths corresponding to this type of rule in $\mathcal{P}$ and Avg Score denotes the average interpretability score of such rules. 0.01 is the threshold value used to select rules with low confidence, which is widely adopted by rule mining methods.
    } 
  \end{table}
  
 \begin{table*}[ht]
  \centering
  \small
  \scalebox{0.8}{
   \begin{tabular}{c|c|c}
    \toprule[1pt]
    \multicolumn{2}{c}{\textbf{Rule:} $\textit{cast member}(X, Y) \leftarrow \textit{producer}(X, A_1) \wedge \textit{spouse}(A_1, Y)$} & \multicolumn{1}{c}{score: 0.0} \\
    \midrule
    1 & (\textit{Veer-Zaara}, \textit{cast member}, \textit{Rani Mukherjee}) $\leftarrow$ (\textit{Veer-Zaara}, \textit{producer}, \textit{Aditya Chopra}) $\wedge$ (\textit{Aditya Chopra}, \textit{spouse}, \textit{Rani Mukherjee}) & score: 0.0 \\
    2 & (\textit{Victor Victoria}, \textit{cast member}, \textit{Julie Andrews}) $\leftarrow$ (\textit{Victor Victoria}, \textit{producer}, \textit{Blake Edwards}) $\wedge$ (\textit{Blake Edwards}, \textit{spouse}, \textit{Julie Andrews}) & score: 0.0\\
    3 & (\textit{The Two Tower}, \textit{cast member}, \textit{Peter Jackson}) $\leftarrow$ (\textit{The Two Tower}, \textit{producer}, \textit{Fran Walsh}) $\wedge$ (\textit{Fran Walsh}, \textit{spouse}, \textit{Peter Jackson}) & score: 0.0\\
    4 & (\textit{10}, \textit{cast member}, \textit{Julie Andrews}) $\leftarrow$ (\textit{10}, \textit{producer}, \textit{Blake Edwards}) $\wedge$  (\textit{Blake Edwards}, \textit{pouse}, \textit{Julie Andrews}) & score: 0.0 \\
    \midrule[1pt]
    \multicolumn{2}{c}{\textbf{Rule:} $\textit{instrument}(X, Y) \leftarrow \textit{occupation}(X, A_1) \wedge \textit{uses}(A_1, Y)$} & \multicolumn{1}{c}{score: 0.9} \\
    \midrule
    1 & (\textit{Sheryl Crow}, \textit{instrument}, \textit{guitar}) $\leftarrow$ (\textit{Sheryl Crow}, \textit{occupation}, \textit{guitarist}) $\wedge$  (\textit{guitarist}, \textit{uses}, \textit{guitar}) & score: 1.0 \\
    2 & (\textit{Tom Waits}, \textit{instrument}, \textit{piano}) $\leftarrow$ (\textit{Tom Waits}, \textit{occupation}, \textit{pianist}) $\wedge$  (\textit{pianist}, \textit{uses}, \textit{piano})& score: 1.0\\
    ... & ... & ... \\
    9 & (\textit{Carly Simon}, \textit{instrument}, \textit{guitar}) $\leftarrow$ (\textit{Carly Simon}, \textit{occupation}, \textit{guitarist}) $\wedge$  (\textit{guitarist}, \textit{uses}, \textit{guitar})& score: 0.5\\
    10 & (\textit{Bruce Hornsby}, \textit{instrument}, \textit{piano}) $\leftarrow$ (\textit{Bruce Hornsby}, \textit{occupation}, \textit{pianist}) $\wedge$  (\textit{pianist}, \textit{uses}, \textit{piano})& score: 1.0\\
    \bottomrule[1pt]
    \end{tabular}
  }
    \caption{Some cases in our labeled dataset (A-benchmark). Every rule corresponds to up to ten paths, each of which has a manually labeled interpretability score. Each rule also has an interpretability score, which is obtained by averaging the interpretability scores of the paths it corresponds to.}
  \label{table:case_study}
 \end{table*}

\paragraph{Manual Annotation.} Because it is difficult for annotators to directly annotate the interpretability score of rules, for every H rule in $\mathcal{F}$, we randomly sample up to ten corresponding real paths in $\mathcal{P}$ for labeling. Here ten is a trade-off between the evaluation effect and the labeling cost. If some rules have less than ten real paths in $\mathcal{P}$, we will take all the paths for labeling. After that, we use the average interpretability scores of the paths corresponding to the rule as the score for this rule. Formally, for a rule $f$, its interpretability score is defined as
\begin{equation}
\label{eq:annotation}
    S(f) = \frac{1}{|P_c(f)|} \sum_{p \in P_c(f)} S^A(p) ,
\end{equation}
where $P_c(f)$ are all sampling paths for rule $f$ and $S^A(p)$ is the interpretability score of the path $p$ obtained by manual annotation. It is worth noting that we cannot annotate the interpretability scores of all paths in $\mathcal{P}$. For the few paths annotated in Equation \ref{eq:annotation}, we can directly obtain their $S(p) = S^A(p)$, while for most of the other paths, we use Equation \ref{eq:approx} to obtain their approximate interpretability scores. For the 15,458 H rules in $\mathcal{F}$, we finally get 102,694 relevant real paths. Each rule corresponds to 6.65 real paths on average.

It is a subjective and difficult thing to give an interpretability score for each path directly. In order to eliminate the influence of subjectivity and make labeling simple, for every true path, we let the annotator choose one of the following three options, namely reasonable, partially reasonable, and unreasonable. These three options correspond to 1, 0.5, and 0 points of the interpretability score, respectively. 
We also try to set the number of options to 2 (reasonable and unreasonable) or 4 (reasonable, most reasonable, few reasonable, and unreasonable), but the final labeling accuracy will decrease. Therefore, we finally adopt the current three-level interpretability score division. 
To further reduces the difficulty of annotation, we use the Graphviz\footnote{\url{https://graphviz.org}} tool to convert the abstract path into a picture. An annotation example is given in Appendix C. 

Table \ref{table:case_study} shows some cases from our labeled dataset, where the interpretability score of a rule is the average of the scores of the paths it samples.

 \subsection{Benchmarks with Mined Rules}
 \label{sec:benchmark_with_mined_rules}
 
 In addition to A-benchmark, we also build benchmarks with mined rules (R-benchmarks). Specifically, we use rule mining methods to mine rules on the training set. These mined rules form a rule set $\mathcal{F}^{*}$ ($\mathcal{F}^{*}$ is $\mathcal{F}^{A}$ for AnyBURL). We can use the confidence of the rule as the interpretability score. But this will introduce another problem, i.e., there is no calibration between the rule's confidence and interpretability score. 
 
 To solve this problem, similar to manual annotation in Section \ref{sec:benchmark_construction}, we need to use 3 classifications (reasonable, partially reasonable, and unreasonable) to label the interpretability score of some rules. We define two thresholds $h_1$ and $h_2$, and the classification $\text{Type}(f)$ of a rule $f$ is defined as:
 \begin{equation}\small
 \label{eq:type}
     \text{Type}(f)=\left\{
     \begin{aligned}
     &\text{unreasonable},\  c(f) < h_1 \\ 
         &\text{partially reasonable},\  h_1 \le c(f) < h_2 \\
         &\text{reasonable},\  c(f) \ge h_2
   \end{aligned}
   \right.
 \end{equation}
 where $c(f)$ is the confidence score of rule $f$. We can regard it as a three-classification task, where the type of prediction is $\text{Type}(f)$, and the golden type is the annotation result. We use Micro-F1 score to find the best $h_1$ and $h_2$, i.e., we search the best $h_1$ and $h_2$ that can get the highest Micro-F1 score. Finally, for every rules $f \in \mathcal{F}$, 
 if $f \notin \mathcal{F}^*$, $S(f) = 0$. Otherwise, we can obtain $\text{Type}(f)$ according to the Equation \ref{eq:type}, and then get the interpretability score $S(f)$. Specifically, unreasonable, partially reasonable and reasonable correspond to 0, 0.5 and 1, respectively.
 
 \paragraph{Discussion} The above two types of benchmarking methods target different situations in terms of accuracy and generalization. On the one hand, manually annotations bring reliable evaluation but are costly and limited in specific datasets (i.e., WD15K). On the other hand, the automatic rule mining method can apply to arbitrary KG completion datasets, while may suffer from inaccurate interpretability scores, as there is no absolute correlation between the confidence and the interpretability score of the rule. 
 
 After obtaining the benchmarks mentioned above, we perform statistical analysis on them. For example, we give the distribution of the interpretability scores and the 20 relations with the highest and lowest interpretability scores. Besides, we also analyze the relation between interpretability scores and confidence scores. Due to space constraints, we put these contents in Appendix B.

 \section{Experiments}

 \subsection{Experimental Setup} 
 
 \paragraph{Models.} We choose two types of multi-hop reasoning models and rule-based reasoning models to evaluate their interpretability.
 For multi-hop reasoning, we use the following five models: MINERVA \cite{MINERVA}, MultiHop \cite{MultiHop}, DIVINE \cite{DIVINE}, R2D2 \cite{R2D2} and RuleGuider \cite{RuleGuider}. For rule-based reasoning, we evaluate the interpretability on the following four models: AMIE+ \cite{AMIE+}, NeuralLP \cite{NeuralLP}, RuLES \cite{RuLES} and AnyBURL \cite{AnyBURL}. We choose them because they are representative models and have well-documented codes for re-implementation. 
 In particular, MultiHop, RuleGuider, and RuLES are all based on knowledge graph embedding models. Referring to the original paper, we also use several variants. The content after the model name represents the embedding model used. For example, MultiHop-\text{\footnotesize{ConvE}} represents the MultiHop model based on ConvE.
 
 \paragraph{Evaluation Protocol.} We target two types of investigation: link prediction and interpretability evaluation. Link prediction mainly tests the performance of the model on KG completion. For every triple $(h, r, t)$ in the test set, we can convert it to a triple query $(h, r, ?)$. Models should give a descending order of the probability that each entity is the correct tail entity.
 We use two evaluation metrics MRR, and Hits@N \cite{ConvE} in experiments. 
 The interpretability evaluation experiment is mainly used to measure the interpretability of reasoning models. Three evaluation metrics, i.e., PR, LI, and GI, are used in this experiment.
 
 \paragraph{Implementation Details.} The distributions of WD15K and FB15K-237 are relatively consistent, so for all baseline models, if the model has parameters provided on FB15K-237, we use these parameters, otherwise we use the default parameters.
 To avoid the influence of beam size on path recall as much as possible, for the RL-based models, we set the beam size to 512, which is four times the commonly used default value of 128.
 For rule mining methods, we set the mining threshold of confidence and head coverage to 0.001. A lower threshold allows the method to mine more rules, which is beneficial to the benchmark introduced in Chapter \ref{sec:benchmark_with_mined_rules}. Besides, we use the Max Aggregation method proposed by AnyBURL to apply the mined rules to rule-based reasoning task. From the original paper of AnyBURL, we can know Max Aggregation can achieve better results than other methods on most datasets.
 
  \begin{table}[t]
  \centering
\setlength\tabcolsep{5.5pt}
  \scalebox{0.68}{
  \begin{tabular}{lcccccccc}
    \toprule
    Model & MRR & @1 & @3 & @10 & PR & LI & GI  \\
  \midrule
  NeuralLP&\textbf{22.9}&19.0&24.6&30.6&10.2&\textbf{80.4}&8.2\\
  AMIE+&-&30.2&40.3&48.7&77.0&{42.1}&{32.4}\\
  RuLES-TransE&-&{44.2}&{56.3}&{67.5}&{92.6}&32.9&30.5\\
  RuLES-HolE&-&43.9&55.2&66.1&91.7&34.0&31.2\\
  AnyBURL&-&45.9&\textbf{58.0}&\textbf{70.4}&\textbf{98.9}&38.4&\textbf{38.0}\\
  \midrule
  MINERVA&42.6&37.5&44.7&51.6&70.7&28.1&19.8\\
  MultiHop-DistMult&\underline{50.3}&\underline{41.8}&\underline{54.8}&\underline{66.8}&89.4&30.6&27.4\\
  MultiHop-ConvE&37.0&24.3&43.2&63.7&91.2&27.0&24.6\\
  MultiHop-TuckER&32.3&20.9&36.3&57.3&\underline{91.3}&28.8&26.3\\
  DIVINE&35.8&27.0&40.3&54.1&67.0&33.0&22.1\\
  R2D2&41.6&36.1&45.9&60.8&7.10&31.5&2.2\\
  RuleGuider-DistMult&{48.0}&38.8&53.0&66.1&89.3&\underline{34.3}&\underline{30.6}\\
  RuleGuider-ConvE&47.8&38.0&53.2&66.7&88.7&30.2&26.8\\
  RuleGuider-TuckER&23.4&13.8&28.0&43.2&74.3&33.3&24.7\\
  \midrule
  Upper Bound&-&-&-&-&99.9&63.4&63.4\\
  \bottomrule
  \end{tabular}
  }
  \caption{Experimental results on WD15K. @1, @3 and @10 denote Hits@1, Hits@3 and Hits@10 metrics, respectively. All metrics are multiplied by 100. The best score of rule-based reasoning models is in \textbf{bold}, and the best score of multi-hop reasoning models is \underline{underlined}. Upper Bound denotes the upper bound interpretability scores given by our A-benchmark.}
  \label{table:main_results}
  \end{table}
 
 \subsection{Results}
 
 Table \ref{table:main_results} show the experimental results on WD15K using A-benchmark. For the link prediction experiment, AnyBURL can achieve comparable or better performance than multi-hop reasoning models. 
 
 In terms of interpretability that this paper is more concerned about, AnyBURL also achieves almost the best results. From a detailed analysis, AnyBURL achieves the highest value of 98.9 in the evaluation metric of PR, which shows that almost all triples in the test set can be found a real path by AnyBURL. For multi-hop reasoning models, there is no big gap with AnyBURL in PR. 
 NeuralLP achieves the highest LI score among all models and is much higher than the other models. But this does not mean that NeuralLP is a highly interpretable model because its PR score is only 10.2. It achieves a high LI score because it only uses rules with high confidence as an explanation. For most triples, it cannot give rules corresponding to a real path. This phenomenon of inconsistency between the evaluation results and the actual interpretability is why we have to elicit GI, which can reflect the overall interpretability of the model.
 
 Among all the models, AnyBURL achieves the highest GI score, which is considerably higher than the second place. RuleGuider is an improved model based on MultiHop. It adds the confidence information of rules (mined by AnyBURL) to the reward, which can improve the interpretability of the model. Judging from the experimental results, such an improvement is indeed effective.
 
 In summary, the current multi-hop reasoning models do have some degree of interpretability. However, compared with the best rule-based reasoning model AnyBURL, multi-hop reasoning models still have some gaps in terms of interpretability and link prediction performance. 
 This reveals that the current multi-hop reasoning models have many shortcomings, and they still lag behind the best rule-based reasoning model. In the future, we should investigate how to better incorporate the mined rules into the multi-hop reasoning model to achieve better performance and interpretability. 
 
 In addition, we also have upper bound scores given by our A-benchmark in Table \ref{table:main_results}, i.e., for each triple, we always choose the path with the highest interpretability score. We can see that Upper Bound is much higher than all models, which indicates that multi-hop reasoning models still have a lot of room for improvement and needs continued research.
 
 \subsection{Manual Annotation vs. Mined Rules}
 
 In order to compare the performance between different benchmarks, all existing models are tested for interpretability on A-benchmark and R-benchmarks. In addition, we have some pruning strategies in the manual annotation. In order to investigate how much these pruning strategies affect the final evaluation performance, we also give the golden interpretability scores of the existing models using a direct evaluation of the sampled paths. Specifically, for each model to be tested, we randomly select 300 paths to the correct tail entity found by the model. After that, we combine and disorganize all the paths and randomly assign them to the annotator to label the interpretability. We use the same three-grade division for labeling as in Section \ref{sec:benchmark_construction}. In addition, each path is assigned to three annotators, and only paths with consistent results from at least two annotators are kept. If a model ends up with less than 300 paths, we continue selection until a sufficient number is reached.
 
 \begin{table}[tb]
  \centering
\setlength\tabcolsep{5.5pt}
  \scalebox{0.7}{
  \begin{tabular}{lcccccccc}
    \toprule
    Model & Manual & AM & RT & RH & AB & Golden  \\
  \midrule
  NeuralLP&8.2&8.6&8.6&8.4&9.2&9.3\\
  AMIE+&32.4&-&17.6&13.0&23.6&26.5\\
  RuLES-TransE&30.5&15.6&-&19.7&22.4&30.4\\
  RuLES-HolE&31.2&15.9&25.4&-&23.1&30.9\\
  AnyBURL&38.0&16.4&21.7&17.1&-&35.9\\
  \midrule
  MINERVA&19.8&0.4&1.8&0.7&5.4&25.3\\
  MultiHop-DistMult&27.4&5.7&5.0&3.2&11.1&29.1\\
  MultiHop-ConvE&24.6&3.2&3.5&1.9&8.9&24.1\\
  MultiHop-TuckER&26.3&3.2&4.1&2.4&10.6&23.9\\
  DIVINE&22.1&3.4&3.6&2.3&11.1&23.4\\
  R2D2&2.2&0.2&0.3&0.2&1.0&2.5\\
  RuleGuider-DistMult&30.6&3.3&2.8&2.1&10.8&30.1\\
  RuleGuider-ConvE&26.8&4.0&3.5&2.8&11.4&29.1\\
  RuleGuider-TuckER&24.7&2.1&1.7&1.2&8.3&26.2\\
  \midrule
  ABS-DIFF-AVG&1.8&18.3&16.7&18.5&11.8&-\\
  \bottomrule
  \end{tabular}
  }
  \vspace{-0.1cm}
  \caption{The interpretability evaluation results using different benchmarks. All numbers in the table are GI scores and are multiplied by 100. Manual denotes benchmark based on manually labeled dataset. AM, RT, RH and AB denote benchmark with rules mined from AMIE+, RuLES-TransE, RuLES-HolE and AnyBURL, respectively. Golden denotes the golden interpretability results. ABS-DIFF-AVG denotes the average of the absolute value between the interpretability score of this benchmark and the golden score.}
  \label{table:vs_results}
  \end{table}
 
 Table \ref{table:vs_results} shows the results of interpretability evaluation with different benchmarks on WD15K. It is worth noting that some data in the table are missing because it is not reasonable to use the rules mined by the rule mining model to evaluate the model itself. 
 We introduce the ABS-DIFF-AVG in the last row of the table, which represents the average of the absolute value between the interpretability score of this benchmark and the golden score. 
 It can measure the gap between these benchmarks and the golden truth.
 From the table, we can learn that A-benchmark can reflect the real interpretability of the model well, which indicates that the approximation and pruning strategies do not have an enormous impact on the final performance. For R-benchmarks, they also differ from each other. The benchmark based on AnyBURL achieve results closer to those of golden. But they also have some common points, i.e., the evaluation is more accurate for rule-based reasoning models and worse for multi-hop reasoning models.
 
 In summary, A-benchmark can accurately evaluate the interpretability of the model. In contrast, the R-benchmarks cannot provide accurate results. However, it has some advantages of its own. For example, it can be applied to other KGs, which has more generality. In addition, it can provide a reference indicator for comparison between different models' interpretability.
 
 \section{Conclusion}
 
 Multi-hop reasoning for KGs has been widely studied in recent years. However, most previous works are conducted on the assumption that the reasoning paths are reasonable. As a result, most of them only care about the model's performance and neglect the evaluation of path interpretability. In this paper, we propose a framework to quantitatively evaluate the interpretability of multi-hop reasoning models. Based on this framework, we annotate a dataset to form a benchmark named BIMR that can accurately evaluate the interpretability. Besides, we also construct benchmarks with generalization ability using mined rules. Experimental results show that our manually-annotated benchmark achieves similar results to the golden truth, indicating that it can be used to evaluate the model's interpretability automatically. Our experimental results show that the rule-based model AnyBURL outperforms the current multi-hop reasoning models in terms of link prediction performance and interpretability, which indicates a possible future research direction, i.e., how to better incorporate rule information into the multi-hop reasoning model to improve the performance and interpretability.
 
 \section*{Acknowledgments}
 
 This work is supported by the NSFC Key Project (U1736204), a grant from the Institute for Guo Qiang, Tsinghua University (2019GQB0003), a grant from Beijing Academy of Artificial Intelligence (BAAI2019ZD0502) and Alibaba. Xin Lv proposed the main idea of the paper and wrote codes. Yixin Cao participated in most of the discussions and made major revisions to the paper. Lei Hou, Juanzi Li and Zhiyuan Liu participated in some of the discussions and made minor revisions to the paper. Yichi Zhang and Zelin Dai performed a proof reading before submission.
 
 \bibliography{emnlp2021}
 \bibliographystyle{acl_natbib}
 
 \clearpage
 
 \appendix
 
 \section{Dataset Construction}
 
 The detailed steps for building WD15K are:
 
 (1) We align entities in FB15K-237 to Wikidata\footnote{We use the 20190506 snapshot of Wikidata.} using Freebase IDs. Almost all entities (98.7\%) in FB15K-237 can be found in Wikidata and these 14,353 entities make up our initial entity set $\mathcal{E}$.
 
 (2) We use every triple $(h, r, t)$ in Wikidata whose $h$ and $t$ are in $\mathcal{E}$ to form a temporary triple set $\mathcal{T}'$. For all relations $r$ in $\mathcal{T}'$, if the number of triples corresponding to $r$ is greater than 5, we save them and finally form the relation set $\mathcal{R}$. 
 
 (3) For every entity that has more than 10 triples with entities in $\mathcal{E}$ through relations in $\mathcal{R}$, we save them as the additional entity set $\mathcal{E}'$. We randomly select some entities in $\mathcal{E}'$ and add them to $\mathcal{E}$, which eventually increases the number of entities in $\mathcal{E}$ to 15,817.
 
 (4) We extract $(h, r, t)$ whose $h, t \in \mathcal{E}$ and $r \in \mathcal{R}$ from Wikidata as our final triple set $\mathcal{T}$.
 
 \begin{figure}[t]
   \centering
   \includegraphics[width=\linewidth]{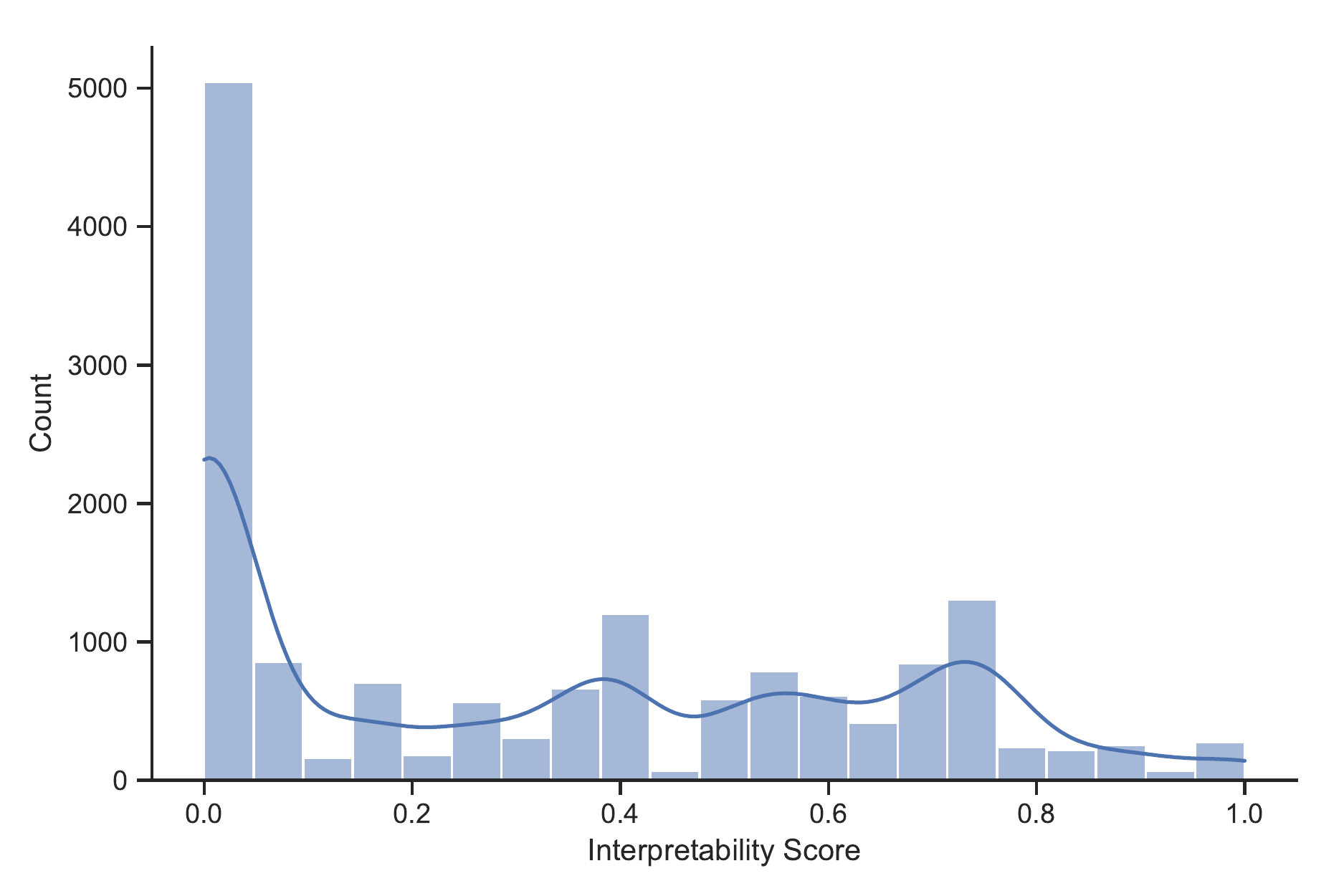}
   \caption{The distribution of interpretability scores for rules in our manually annotated dataset.}
   \label{fig:ours_statistics}
 \end{figure}
 
 \section{Statistics}
 
 \subsection{Manual Annotation Dataset}
 
\begin{figure}[t]
   \centering
   \includegraphics[width=\linewidth]{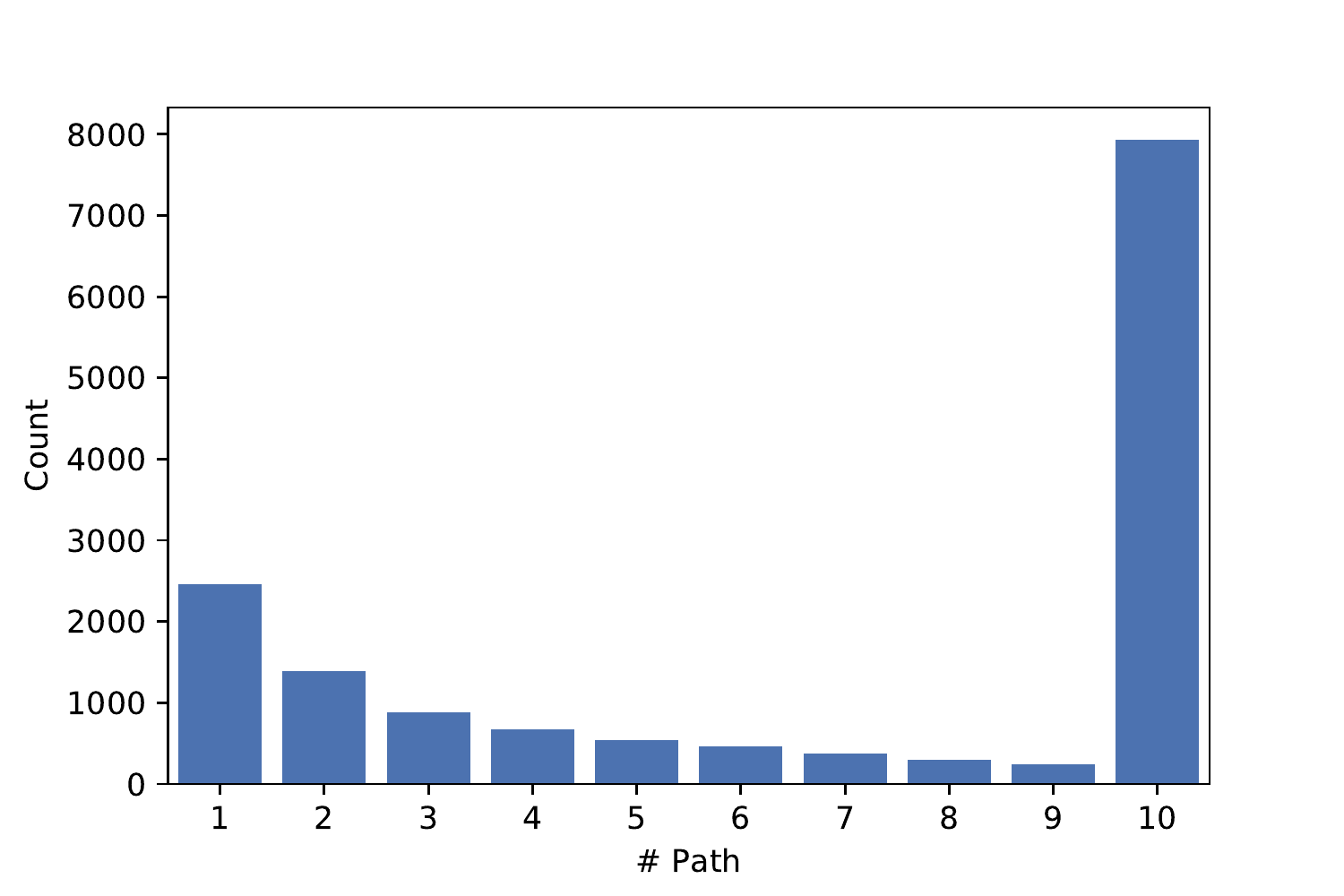}
   \caption{Distribution of the number of paths sampled by each rule in our annotation process.}
   \label{fig:path}
 \end{figure}

 In this section, we give some statistical information about our manually annotated dataset. Specifically, our dataset contains a total of 15,458 N rules with interpretability score, and they are obtained from the interpretability scores of 102,694 paths. Figure \ref{fig:ours_statistics} shows the distribution of the interpretability scores of these rules. Figure \ref{fig:path} represents the distribution of the number of paths corresponding to each rule. 
 
\begin{figure*}[t]
   \centering
   \includegraphics[width=160mm]{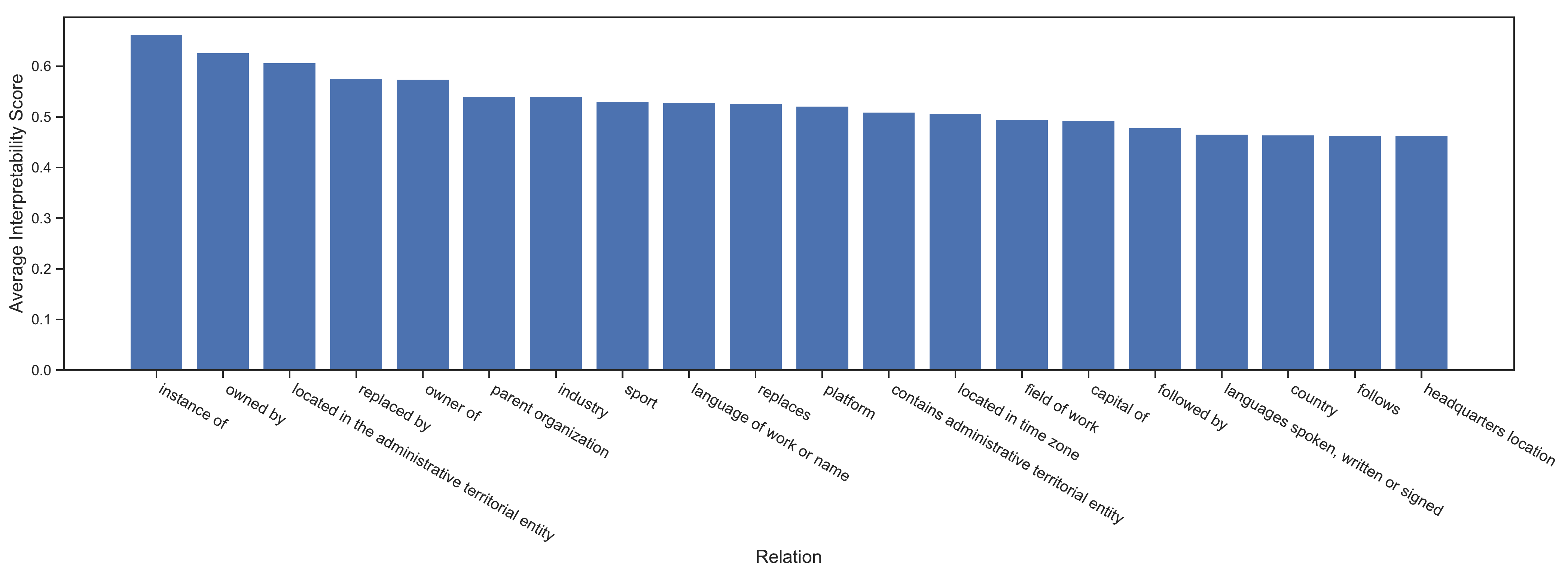}
   \caption{The 20 relations with the highest average interpretability scores in our annotation data and their average scores.}
   \label{fig:max_20}
 \end{figure*}

\begin{figure*}[t]
   \centering
   \includegraphics[width=160mm]{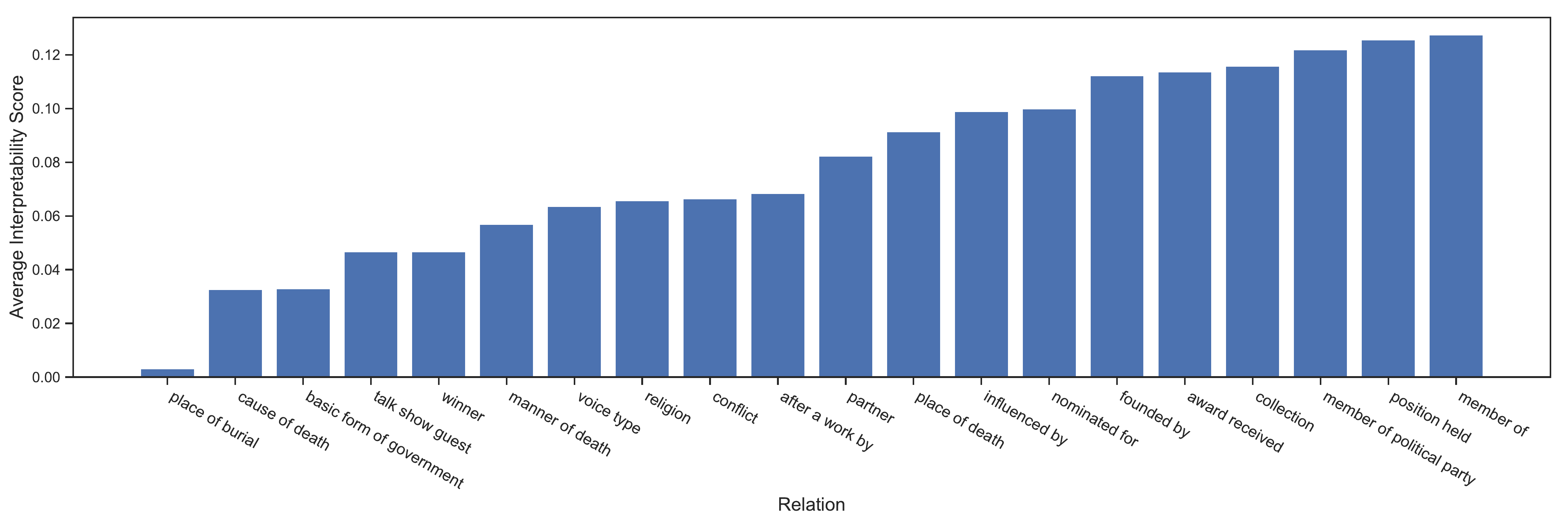}
   \caption{The 20 relations with the lowest average interpretability scores in our annotation data and their average scores.}
   \label{fig:min_20}
 \end{figure*}

 The reasonability of each relation in the knowledge graph is different. For example, the relation \textit{cause of death} is difficult to obtain by a reasonable reasoning path. We calculate the average value of the interpretability score of the rules corresponding to each relation. For example, for the relation \textit{cause of death}, we find all the rules that can derive the relation \textit{cause of death} (e.g., $\textit{cause of death}(X, Y) \leftarrow \textit{spouse}(X, A_1) \wedge \textit{cause of death}(A_1, B)$). The average of the interpretability of these rules is the value we need. To avoid the effect of long-tail relations, we only consider relations with at least 10 rules. We give the 20 relations with the highest and lowest average interpretability scores in Figures \ref{fig:max_20} and \ref{fig:min_20}, respectively.
 
 \subsection{Mined Rules}
 
\begin{figure*}[t]
   \centering
   \includegraphics[width=160mm]{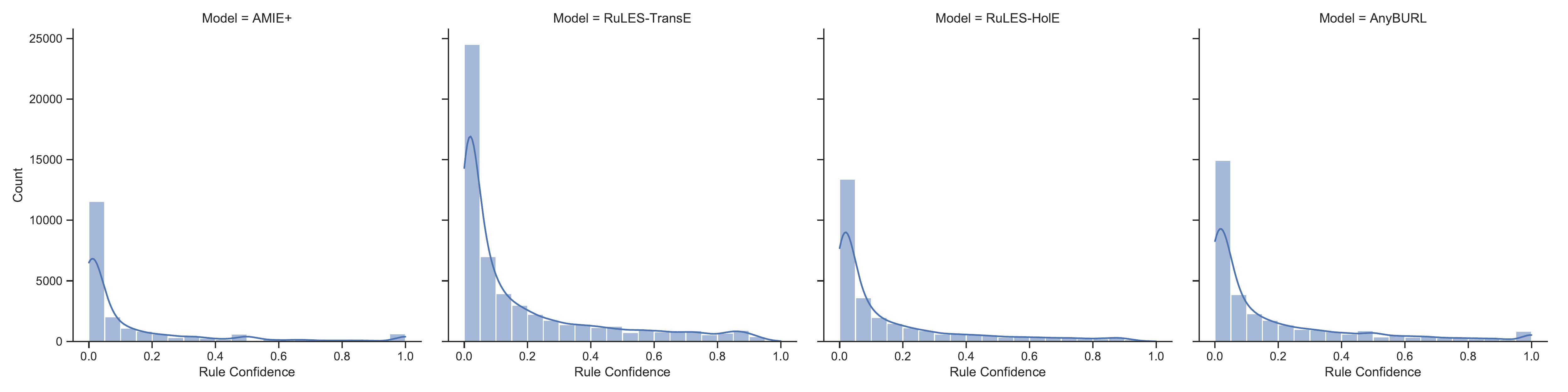}
   \caption{Distribution of confidence scores of rules mined by different rule mining methods on the training set of WD15K.}
   \label{fig:rules_statistics}
 \end{figure*}

\begin{figure*}[t]
   \centering
   \includegraphics[width=160mm]{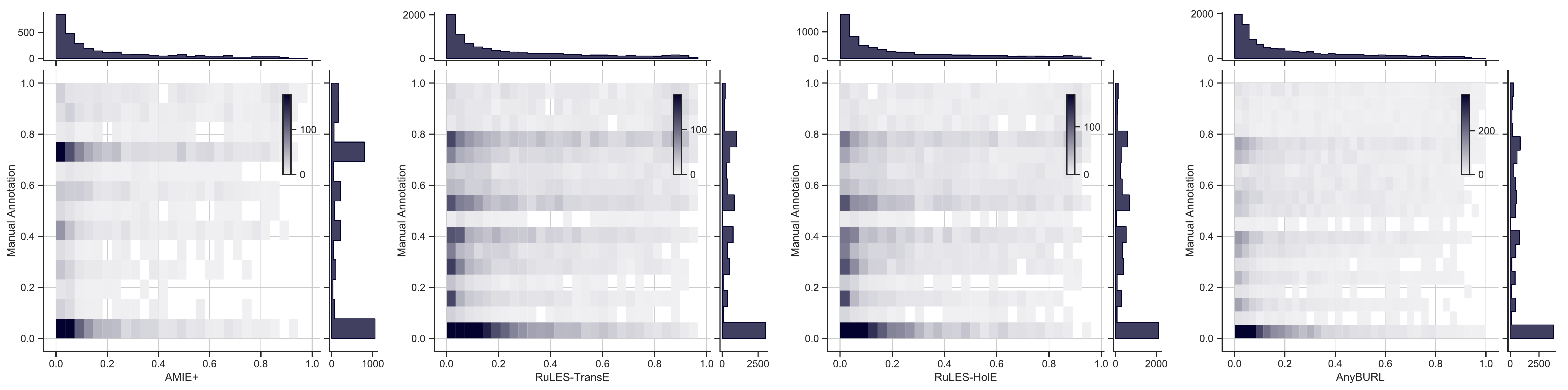}
   \caption{The joint distribution between the confidence score of the rules mined by the rule mining tool and the interpretability score in our labeled data. Specifically, the horizontal coordinate is the confidence score of the rules mined by each rule mining tool, and the vertical coordinate is the interpretability score in our labeled data.}
   \label{fig:rules}
 \end{figure*}

 For the four rule mining models, AMIE+, RuLES-TransE, RuLES-HolE and AnyBURL, we use them to perform rule mining on the training set of WD15K. Figure \ref{fig:rules_statistics} shows the distribution of the confidence scores of the rules mined by these models. In addition, we give the joint distribution between the rule confidence of these models and our manually labeled interpretability scores in Figure \ref{fig:rules}. The joint distribution only calculates rules that appear in the set of 15,458 N rules we labeled.
 
 \section{Annotation Example}
 
\begin{figure*}[t]
   \centering
   \includegraphics[width=160mm]{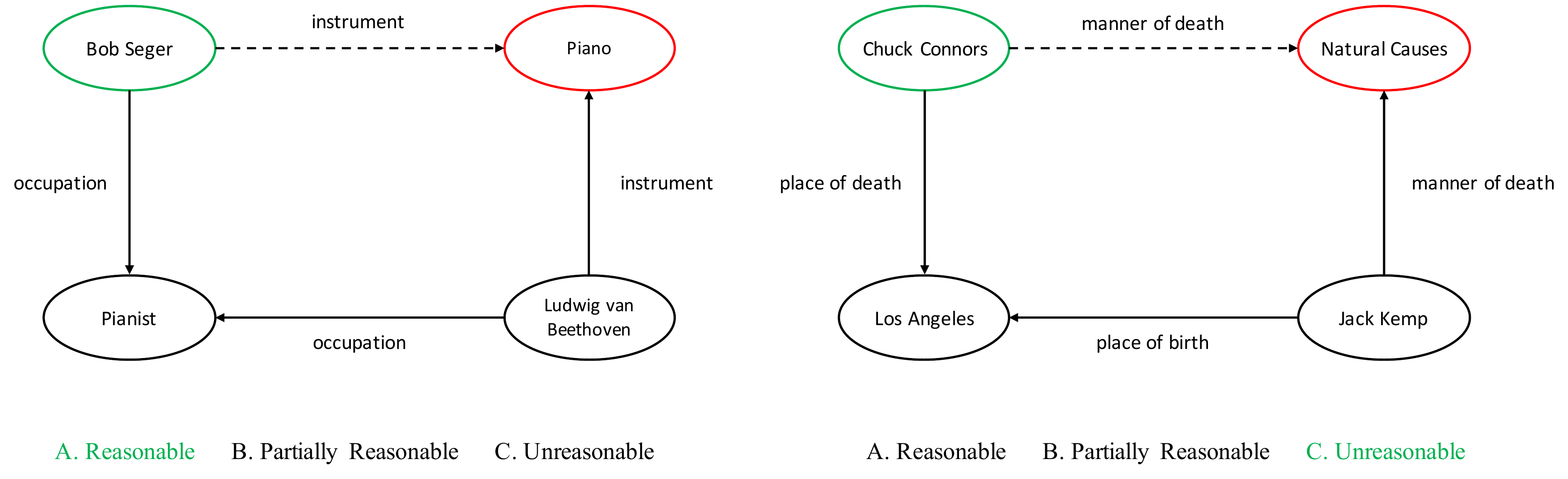}
   \caption{Some annotation examples in our annotation process, the annotator needs to judge whether this reasoning path is reasonable, and choose the correct one among the three options. The green option is the correct answer.}
   \label{fig:example}
 \end{figure*}

 We give some annotation examples in our annotation process in Figure \ref{fig:example}.
 
 \section{Computing Infrastructure}
 
 Our experiments are run on the server with the following configurations:
 
 \begin{itemize}
   \item OS: Ubuntu 16.04.6 LTS
   \item CPU: Intel(R) Xeon(R) CPU E5-2680 v4 @ 2.40GHz
   \item GPU: GeForce RTX 2080 Ti
 \end{itemize}
 
 \end{document}